\documentclass[letterpaper, twocolumn, oneside, 10pt]{article}
\usepackage{usenix}
\PassOptionsToPackage{hyphens}{url}\usepackage{hyperref}
\usepackage{graphicx}
\usepackage{color}
\usepackage[sort&compress,square,numbers]{natbib}
\usepackage{multirow}
\usepackage{amsmath}
\usepackage{booktabs}
\usepackage{hyperref}
\usepackage{fullpage}
\usepackage[font=small,skip=0pt]{caption}

\usepackage{titling}
\usepackage{hyperref}

\begin{document}
\setlength{\droptitle}{-4em}   
\title{MagNet and ``Efficient Defenses Against Adversarial Attacks'' \\ are Not Robust to Adversarial Examples}
\date{}
\author{Nicholas Carlini \qquad David Wagner \\ University of California, Berkeley}
\maketitle

\section*{Abstract}
MagNet and ``Efficient Defenses...'' were recently proposed as a defense
to adversarial examples.
We find that we can construct adversarial examples that
defeat these defenses with only
a slight increase in distortion.

\section{Introduction}

It is an open question how to
train neural networks so they will be robust to adversarial examples
\cite{szegedy2013intriguing}.
Recently, three defenses have been proposed to make neural networks
robust to adversarial examples:
\begin{itemize}
\item MagNet \cite{meng2017magnet}
was proposed as an approach
to make
neural networks robust against adversarial examples through two
complementary approaches: adversarial examples near the data manifold are
\emph{reformed} to lie on the data manifold that are classified
correctly, whereas adversarial examples far away from the data manifold
are \emph{detected} and rejected before classification.
MagNet does not argue robustness in the white-box setting; rather, the
authors argue that MagNet is robust in the grey-box setting where the
adversary is aware the defense is in place, knows the parameters of
the base classifier, but \emph{not} the parameters of the defense.
\item An efficient defense
\cite{zantedeschi2017efficient}
was proposed to make
neural networks more robust against adversarial examples by performing
Gaussian data augmentation during training, and using the BReLU
activation function. The authors do not claim perfect security,
but claim this makes attacks visually detectable.
\item Adversarial Perturbation Elimination GAN (APE-GAN) \cite{apegan}
  is similar to MagNet, only adversarial examples are projected
  onto the data manifold using a Generative Adversarial
  Network (GAN) \cite{goodfellow2014generative}. We did not set out
  to bypass this defense, but found it to be very similar to MagNet
  and so we analyze it too.
\end{itemize}

In this short paper, we demonstrate these three defenses are not
effective on the MNIST \cite{lecun1998gradient} and
CIFAR-10 \cite{krizhevsky2009learning} datasets.
We show that we are able to bypass MagNet
with greater than $99\%$ success, and the latter two with $100\%$,
with only a slight increase in distortion.

We defeat MagNet by making use of
the \emph{transferability} \cite{szegedy2013intriguing}
property of adversarial examples: the
adversary trains their own copy of the defense, constructs
adversarial examples on their model, and supplies these adversarial
examples to the defender. It turns out that these examples will
also fool the defender's model.

We defeat ``Efficient Defenses Against Adversarial Attack'' and APE-GAN
by showing that existing attack can defeat them
with $100\%$ success without modification.
Adversarial examples are not more visually detectable than
an undefended network.

\section{Background}

We assume familiarity with neural networks \cite{lecun1998gradient}, adversarial examples
\cite{szegedy2013intriguing}, transferability \cite{liu2016delving},
generating strong attacks against
adversarial examples \cite{carlini2016towards} and
MagNet \cite{meng2017magnet}.
We briefly review the key details and notation.

\bigskip \noindent
\textbf{Notation}
Let $F(x) = y$ be a neural network used for classification outputting a
probability distribution.
Call the second-to-last layer (the layer before the
the softmax layer) $Z$, so that $F(x) = \text{softmax}(Z(x))$.
Each output $y_i$ corresponds to the
predicted probability that the object $x$ is labelled as class $i$. Let
$C(x) = \arg \max_i \ F(x_i)$ correspond to the classification
of $x$ on $F$. In this paper we are concerned with neural
networks used to classify images (on MNIST and CIFAR-10).

\bigskip \noindent
\textbf{Adversarial examples} \cite{szegedy2013intriguing}
are instances $x'$ that are very close to a
normal instance $x$ with respect to some distance metric ($L_2$ distance,
in this paper),
but where $C(x') = t$ for any target $t$ chosen by the adversary.

We generate adversarial examples with Carlini and Wagner's $L_2$ attack algorithm
\cite{carlini2016towards}. Specifically, we solve
\[ \text{minimize } \; \|x' -x\|_2^2 + c \cdot \ell(x') \]
where the loss function $\ell$ is defined as
\[ \ell(x') = \max(\max \{ Z(x')_i : i \ne t\} - Z(x')_t, 0). \]
and the constant $c$ is chosen via binary search.

Perhaps the most surprising property of adversarial examples is
\emph{transferability} \cite{szegedy2013intriguing}: given two different models
that perform the same task, trained on different datasets, adversarial
examples constructed to fool one one model often fool the other as well.
This effect can be amplified \cite{liu2016delving} by constructing the
adversarial examples to fool multiple local models before applying it
to the target model.

\bigskip \noindent
\textbf{MagNet} \cite{meng2017magnet}
was recently proposed as a defense to adversarial examples under the
\emph{grey box} threat model: here, the adversary is aware that MagNet is
being used to secure a given classifier, as well as the parameters of the
classifier, but is not aware of the parameters of the defensive models.
MagNet has two defense components:
\begin{enumerate}
\item The \emph{detector} examines inputs and decides if they are
  adversarial or benign. There can be multiple detectors, in which case
  if any report ``adversarial'' the input is rejected.
\item The \emph{reformer} takes inputs classified as benign by the
  detector(s) and reforms them to remove any remaining minor adversarial
  nature to the instances. The defender trains multiple reformers and
  selects one at random, to prevent the adversary from learning its
  parameters.
\end{enumerate}
The authors argue these defenses complement each other: the detector should
be able to detect any adversarial examples with large distortion and are
far away from the true data manifold. The reformer takes instances which
are close to the data manifold and maps them directly on to the
manifold to remove any minor adversarial perturbations  that may not have been
detected.

The specific instantiation of the detector and reformer suggested
by the authors are based
on auto-encoders. Given an auto-encoder $AE$, to detect if an input is adversarial,
compare the mean squared error $\|x-AE(x)\|_2$ with a threshold
$\tau$ chosen to minimize the false positive rate. On CIFAR, the authors
suggest one further detector: compute the Jensen-Shannon divergence
$JSD(F(x)/T,F(AE(x))/T)$ with temperature $T$ (set to $10$ and $40$),
and report ``adversarial'' if the difference is greater than a different
threshold $\tau'$, also chosen to minimize the false positive rate.

Similarly, to reform an adversarial example run it through the
auto-encoder and let $x \gets AE(x)$: since the auto-encoder is trained
to reconstruct typical instances, this operation projects
it back to the data manifold.

\bigskip \noindent
\textbf{Efficient Defenses Against Adversarial Attacks} \cite{zantedeschi2017efficient}
works by making two modifications to standard neural networks.
First, the authors propose use of the Bounded ReLU activation function, defined as
$BReLU(x) = \text{min}(\text{max}(x,0),1)$ instead of standard ReLU \cite{nair2010rectified}
which is unbounded above.
Second, instead of training on the standard training data
$\{(x_i, y_i)\}_{i=1}^n$ they train on
$\{(x_i + N_i, y_i)\}_{i=1}^n$ where $N_i \sim \mathcal{N}(0, \sigma^2)$ is chosen
fresh for each training instance. On MNIST, $\sigma=0.3$; for CIFAR, $\sigma=0.05$.
The authors claim that despite training on noise, it is successful on \cite{carlini2016towards}.

\bigskip \noindent
\textbf{APE-GAN} \cite{apegan}
works by constructing a pre-processing network $G(\cdot)$
trained to project both normal instances and adversarial examples back to the
data manifold as done in MagNet. The network $G$ is trained with a GAN instead
of an auto-encoder. Note that unlike a standard GAN which takes as input a noise
vector and must produce an output image, the generator in APE-GAN takes in an
adversarial example and must make it appear non-adversarial. During training,
the authors train on adversarial examples generated with the Fast Gradient
Sign algorithm \cite{goodfellow2014explaining}; despite this,
the authors claim robustness on a
wide range of attacks (including \cite{carlini2016towards}).

\bigskip \noindent
\textbf{Defense Models.}
We take the MagNet implementation from the authors' open-source code
\footnote{\url{https://github.com/Trevillie/MagNet}} and train our own models.
Since the provided code does not include an implementation of the CIFAR
defense and classifier, we implement it as described in the paper.

We were unable to obtain source code for ``Efficient Defenses...''. We therefore
re-implement the proposed defense based on the description in the paper.
We take the APE-GAN implementation from the authors open-source code
\footnote{\url{https://github.com/shenqixiaojiang/APE-GAN}}.

We use the C\&W attack given by the authors.
We also release our code
\footnote{\url{https://github.com/carlini/MagNet}}
\footnote{\url{https://github.com/carlini/breaking_efficient_defenses}}
\footnote{\url{https://github.com/carlini/APE-GAN}}.

\section{Breaking MagNet}

We demonstrate MagNet is not robust by generating
transferable adversarial examples with Carlini and Wagner's $L_2$ attack.
To do this, we (as the adversary) train $32$ local auto-encoders to use
as detectors and reformers (different than the 16 the defender will use).
To generate adversarial examples, we
modify the optimization formulation to the following, where reformers are
denoted $R_j$ and the detectors $D_j$:
\[ \text{minimize } \; \|x' -x\|_2^2 + c \cdot \ell_c(x') + d \cdot \ell_d(x') \]
where the \emph{classifier} loss function $\ell_c$ is defined as
\[ \ell_c(x') = \sum\limits_j \max(\max\limits_{i \ne t} Z(R_j(x'))_i - Z(R_j(x'))_t, -\kappa) \]
and the \emph{detector} loss function $\ell_d$ is defined as 
\[ \ell_d(x') = \sum\limits_j \max(D_j(x') - \tau_j, 0). \]
Here, $c$ and $d$ are chosen via binary search simultaneously.
The threshold $\tau_j$ is chosen such that $1\%$ of validation images
are detected as adversarial. (The true defender sets their remote $\tau$ to
reject only $0.1\%$
of instances \cite{meng2017magnet}: ours is more strict to
help the attack construct adversarial
examples that fool the remote detectors more often.)
We set the confidence $\kappa=1$ so that the adversarial examples are more
likely to transfer.

\begin{table}
\centering
\begin{tabular}{llrr}
  \toprule
  Dataset & Model & Success & Distortion ($L_2$) \\
  \midrule
  \multirow{2}{*}{MNIST} & Unsecured & 100\% & 1.64\\
  & MagNet & 99\% & 2.25\\
  \midrule
  \multirow{2}{*}{CIFAR} & Unsecured & 100\% & 0.30\\
  & MagNet & 100\% & 0.45\\
 \bottomrule
\end{tabular}
\vskip 0.1in
\caption{The success rate of our attack on MagNet.
  The last column shows the
  mean distance to the nearest targeted adversarial example, across
  the first 1000 test instances, with the target chosen uniformly at random
  from the incorrect classes.}
  \label{tbl:models_magnet}
\end{table}

We attack by performing $10000$ iterations of gradient descent with a learning
rate of $10^{-2}$.
We did not perform hyperparameter search (e.g., picking $32$ auto-encoders,
$\kappa=1$, $\tau_j=0.01$);
improved search over these parameters would yield lower distortion adversarial
examples.

Figures~1 and 2 contain images of targeted adversarial examples on
on the secured network, and Table~\ref{tbl:models_magnet} the mean distortion required across the
first 1000 instances of the test set with targets chosen uniformly at random
among the incorrect classes.

\begin{figure}
  \includegraphics[scale=.15]{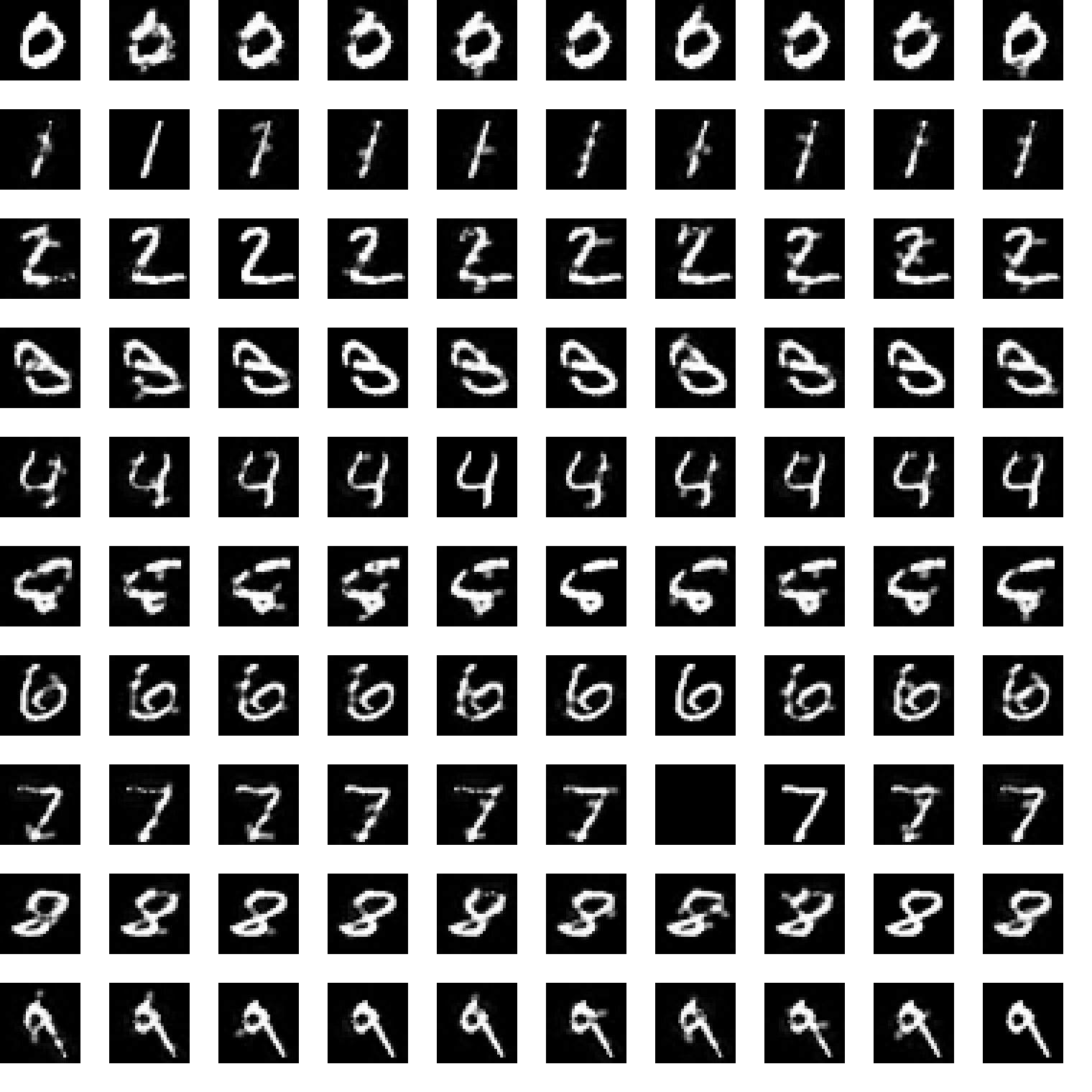}
  \caption{MagNet targeted adversarial examples for each source/target pair of images on MNIST.
  We achieve a $99\%$ grey-box success (the $7 \to 6$ attack failed to transfer).}
\end{figure}
\begin{figure}
  \includegraphics[scale=.1353]{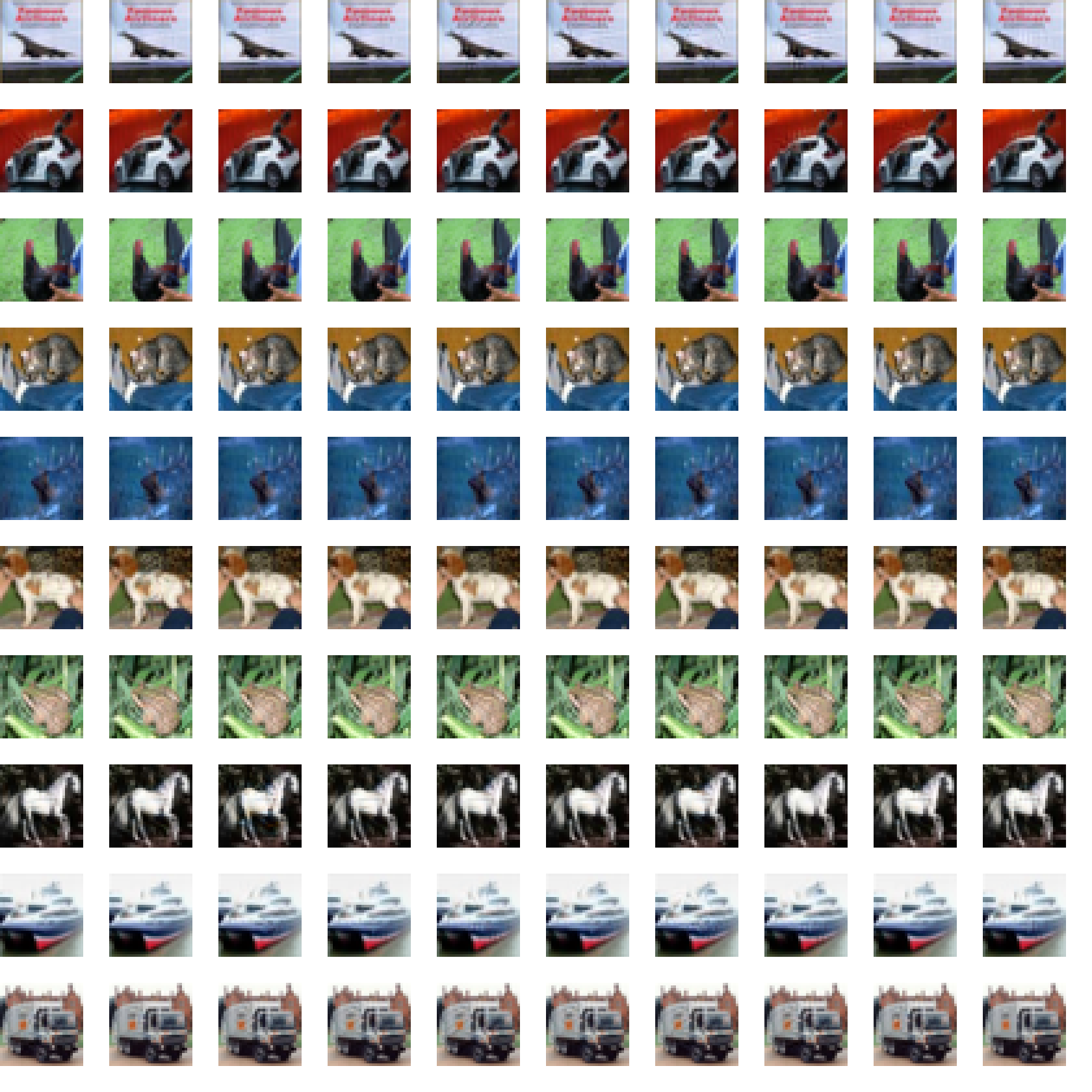}
  \caption{MagNet targeted adversarial examples for each source/target pair of images on CIFAR.
  We achieve a $100\%$ grey-box success.}
\end{figure}

\newpage
\section{Breaking ``Efficient Defenses...''}

We demonstrate this defense is not robust by generating
adversarial examples with Carlini and Wagner's $L_2$ attack.
We do nothing
more than apply the attack to the defended network.

Figure~3 contains images of adversarial examples
on the secured network, and Table~\ref{tbl:models_vnr} the mean distortion required across the
first 1000 instances of the test set with targets chosen at random
among the incorrect classes.

\begin{table}
\centering
\begin{tabular}{llrr}
  \toprule
  Dataset & Model & Distortion ($L_2$) \\
  \midrule
  \multirow{4}{*}{MNIST} & Unsecured & 2.04\\
  & BReLU & 2.14\\
  & Gaussian Noise & 2.66\\
  & Gaussian Noise + BReLU & 2.58\\
  \midrule
  \multirow{4}{*}{CIFAR} & Unsecured & 0.56\\
  & BReLU & 0.58\\
  & Gaussian Noise & 0.66\\
  & Gaussian Noise + BReLU & 0.67\\
 \bottomrule
\end{tabular}
\vskip 0.1in
\caption{Neither adding Gaussian data augmentation during
  training nor using the BReLU activation significantly increases
  robustness to adversarial examples on the MNIST or CIFAR-10 datasets;
  success rate is always $100\%$.}
  \label{tbl:models_vnr}
\end{table}

On MNIST, the full defense increases mean distance to the nearest adversarial
example by $30\%$, and on CIFAR by $20\%$. This is in contrast with other
forms of retraining, such as adversarial retraining
\cite{madry2017towards}, which increase distortion
by a significantly larger amount.
Interestingly, we find that BReLU provides some increase in distortion when
trained without Gaussian augmentation, but when trained with it, does not
help.

\section{Breaking APE-GAN}

We demonstrate APE-GAN is not robust by generating
adversarial examples with Carlini and Wagner's $L_2$ attack.
We do nothing
more than apply the attack to defended network. That is, we change the loss
function to account for the fact that the manifold-projection is done before
classification. Specifically, we let
\[ \ell(x') = \max(\max \{ Z(G(x'))_i : i \ne t\} - Z(G(x'))_t, 0) \]
and solve the same minimization formulation.

Figure~4 contains images of adversarial examples
on APE-GAN, and Table~\ref{tbl:models_apegan} the mean distortion required across the
first 1000 instances of the test set with targets chosen at random
among the incorrect classes.

\bigskip \noindent
\textbf{Investigating APE-GAN's Failure.}
Why are we able to fool APE-GAN? We compare (a) the mean
distance between the original inputs and the adversarial
examples, and (b) the mean distance between the original inputs and
the recovered adversarial examples. We find that the recovered
adversarial examples are \emph{less similar} to the original
than the adversarial examples. Specifically, the mean distortion
between the adversarial examples and the original instances is $4.3$,
whereas the mean distortion between the recovered instances
and original instances is $5.8$.

This indicates that what our adversarial examples have done
is fool the generator $G$ into giving reconstructions that
are even less similar from the original than the adversarial
example. This effect can be observed in Figure~4: faint lines
introduced become more pronounced after reconstruction.

\begin{table}[t]
\centering
\begin{tabular}{llrr}
  \toprule
  Dataset & Model & Success & Distortion ($L_2$) \\
  \midrule
  \multirow{2}{*}{MNIST} & Unsecured & $100\%$ & 2.04\\
  & APE-GAN & $100\%$ & 2.17\\
  \midrule
  \multirow{2}{*}{CIFAR} & Unsecured & $100\%$ & 0.43\\
  & APE-GAN & $100\%$ & 0.72\\
 \bottomrule
\end{tabular}
\vskip 0.1in
\caption{APE-GAN does not significantly increase robustness to
  adversarial examples on the MNIST or CIFAR-10 datasets.}
  \label{tbl:models_apegan}
\end{table}

\section{Conclusion}
As this short paper demonstrates, MagNet is not robust to transferable
adversarial examples, and  combining Gaussian data augmentation and
BReLU activations does
not significantly increase the robustness of a neural network
against strong iterative attacks.
Surprisingly, we found that while all three defenses take different
approaches to increasing the robustness against adversarial examples,
they all give approximately the same increase in robustness ($\sim30\%$).

We recommend that researchers who
propose defenses attempt adaptive white-box attacks against their schemes
before claiming robustness.
Or, if arguing security in the grey-box threat model, we recommend
researchers generate adversarial examples targeting \emph{the specific defense}
by using a copy of that defense as the source model.
Just because the adversary is not aware of the exact parameters, does not
mean the best that can be done is to transfer from an unsecured model:
as we have shown here, transfering from a local copy of the model improves
the attack success rate.

\begin{figure}
  \includegraphics[scale=.105]{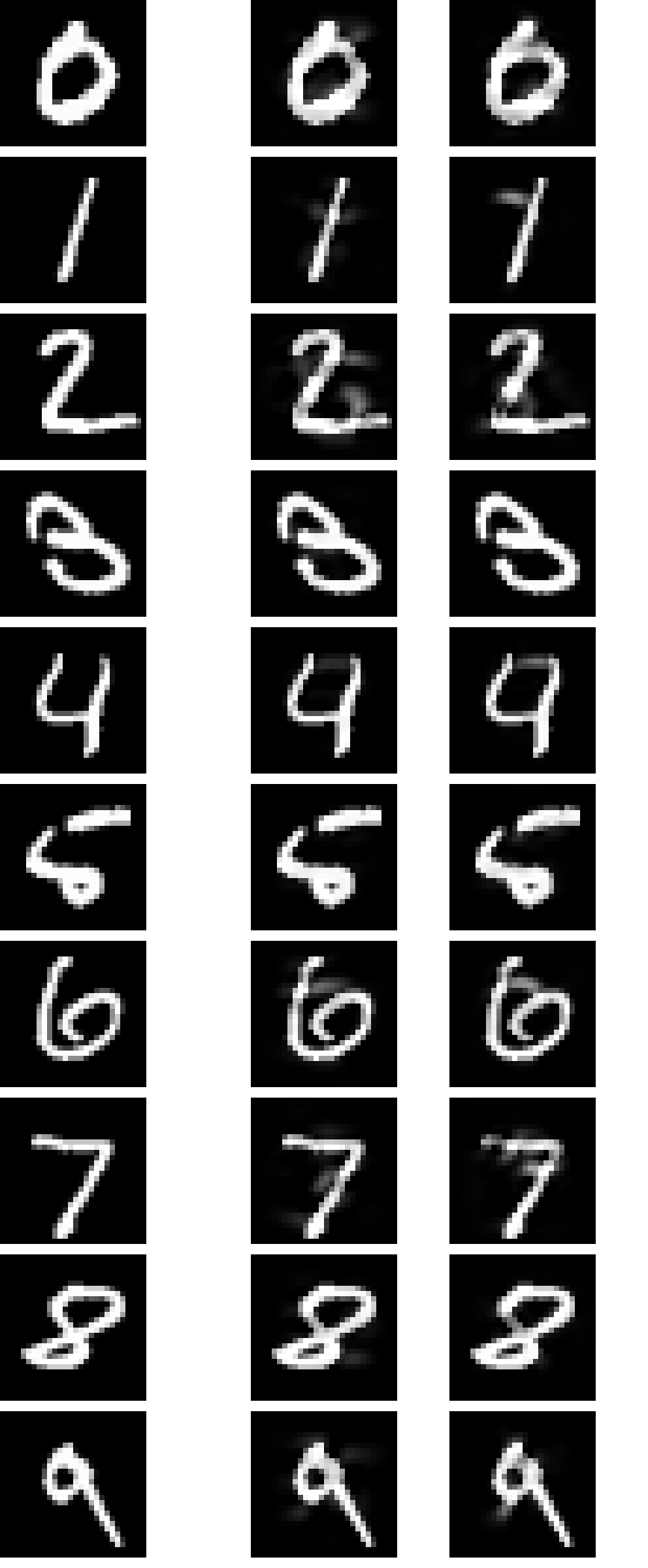}
  \hspace{1em}
  \includegraphics[scale=.105]{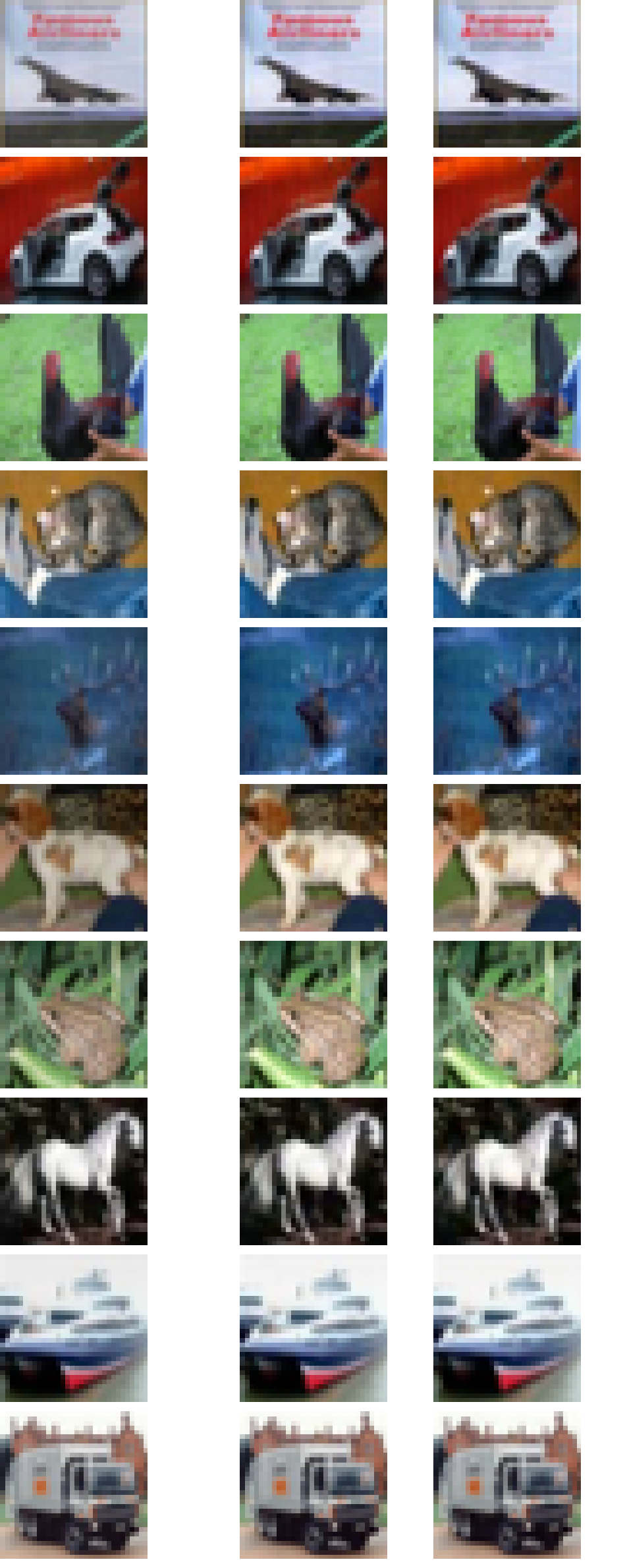} \\
  \emph{(a) \hspace{2.5em} (b) \hspace{1.6em} (c) \hspace{2.9em}  (a) \hspace{2.3em} (b) \hspace{1.5em} (c)}
  \caption{Attacks on ``Efficient Defenses...'' on  MNIST and CIFAR-10: \emph{(a)} original reference image; \emph{(b)} adversarial example on the defense with only BReLU; \emph{(c)} adversarial example on the complete defense with Gaussian noise and BReLU.}
\end{figure}
\begin{figure}
  \includegraphics[scale=.105]{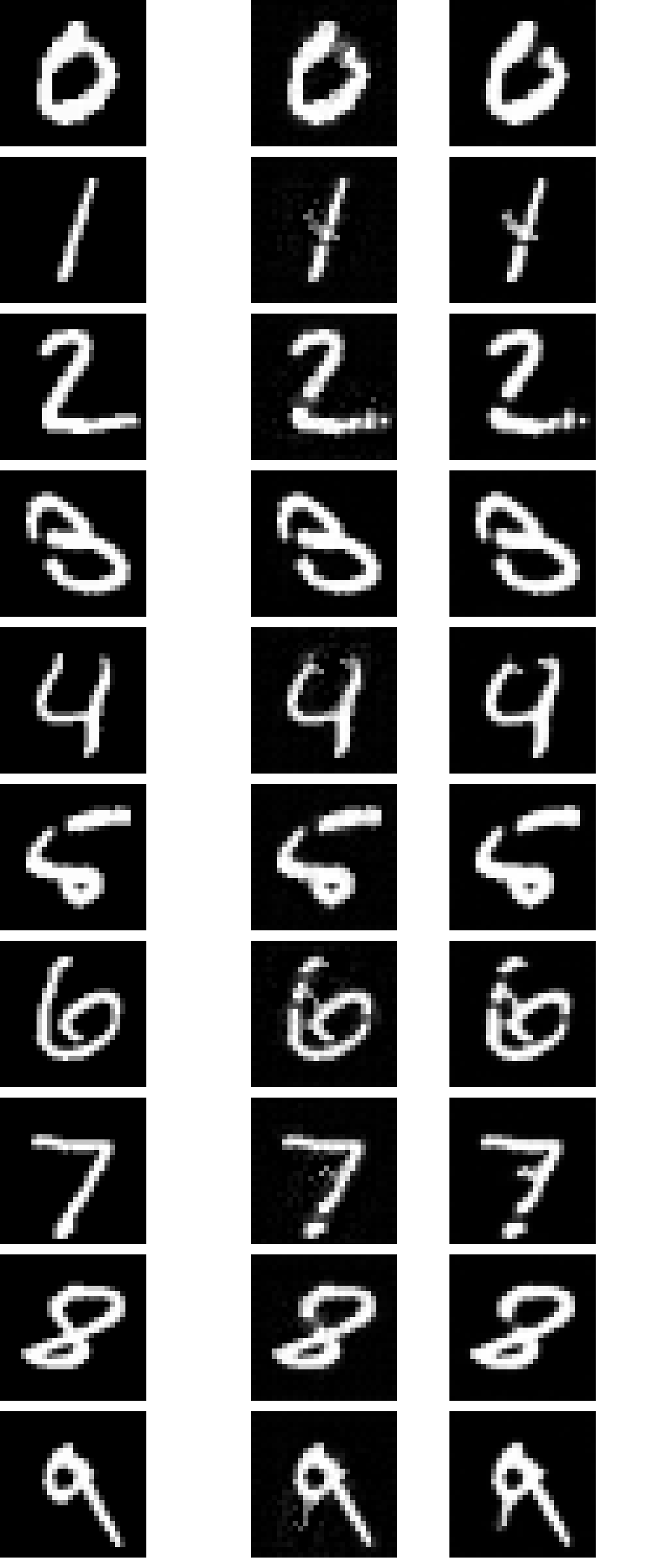}
  \hspace{1em}
  \includegraphics[scale=.105]{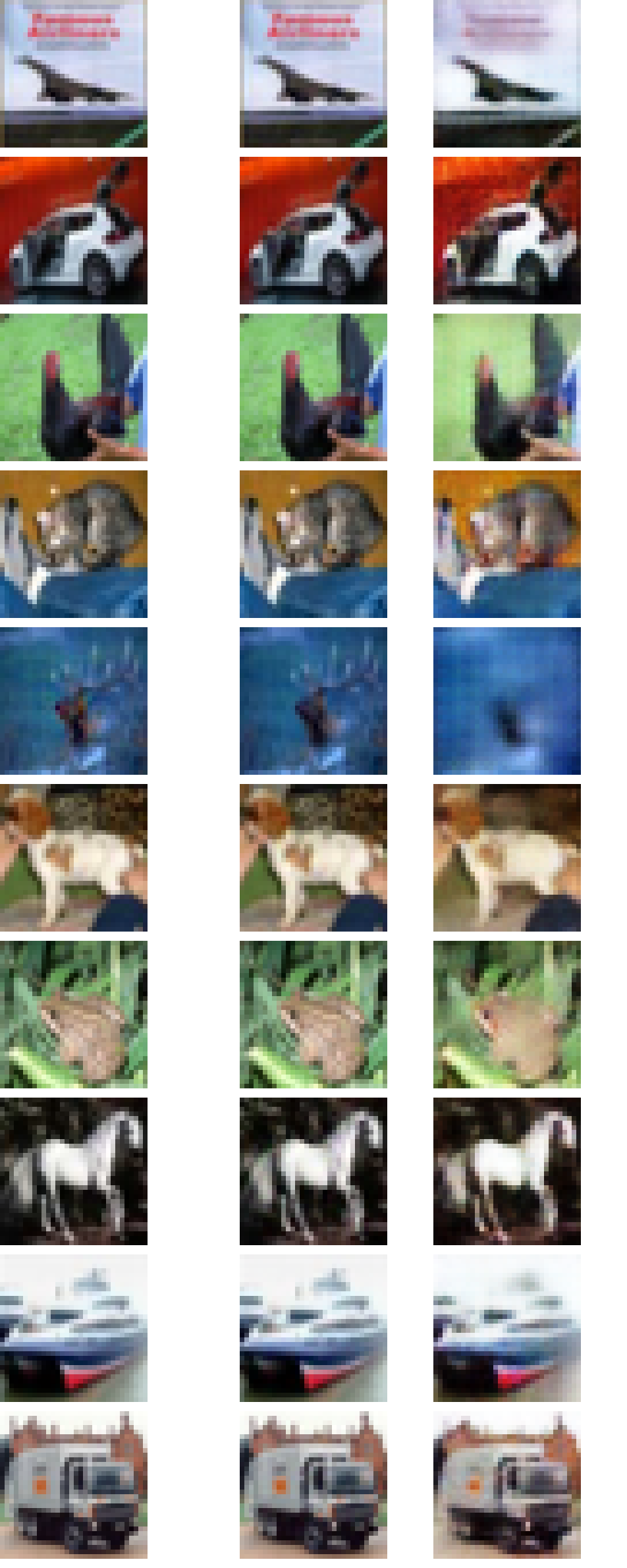} \\
  \emph{(a) \hspace{2.5em} (b) \hspace{1.6em} (c) \hspace{2.9em}  (a) \hspace{2.3em} (b) \hspace{1.5em} (c)}
  \caption{Attacks on APE-GAN on MNIST and CIFAR-10: \emph{(a)} original reference image; \emph{(b)} adversarial example on APE-GAN;
    \emph{(c)} reconstructed adversarial example. }
\end{figure}

{\footnotesize
\bibliographystyle{abbrvnat}
\bibliography{paper}
}

\end{document}